\documentclass[runningheads]{llncs}
\usepackage{graphicx}
\usepackage{multirow}
\usepackage{amsfonts}
\usepackage{xspace}
\usepackage{subfigure}
\usepackage{enumitem}
\usepackage{xcolor}
\usepackage{cite}
\usepackage[misc]{ifsym}
\definecolor{citecolor}{HTML}{0071bc}
\usepackage[pagebackref,breaklinks=true,letterpaper=true,colorlinks,citecolor=citecolor,bookmarks=false]{hyperref}

\begin{document}

\title{Graph-level Anomaly Detection via Hierarchical Memory Networks} 

\author{Chaoxi Niu \inst{1} \and Guansong Pang(\Letter) \inst{2}\thanks{G. Pang is the corresponding author.} \and Ling Chen \inst{1}}

\authorrunning{C. Niu et al.}

\institute{AAII, University of Technology Sydney, Sydney, Australia \\
\email{Chaoxi.Niu@student.uts.edu.au, Ling.Chen@uts.edu.au} \and Singapore Management University, Singapore \\
\email{pangguansong@gmail.com}}

\maketitle

\begin{abstract}
Graph-level anomaly detection aims to identify abnormal graphs that exhibit deviant structures and node attributes compared to the majority in a graph set. One primary challenge is to learn normal patterns manifested in both fine-grained and holistic views of graphs for identifying graphs that are abnormal in part or in whole. To tackle this challenge, we propose a novel approach called Hierarchical Memory Networks (HimNet), which learns hierarchical memory modules -- node and graph memory modules -- via a graph autoencoder network architecture. The node-level memory module is trained to model fine-grained, internal graph interactions among nodes for detecting locally abnormal graphs, while the graph-level memory module is dedicated to the learning of holistic normal patterns for detecting globally abnormal graphs. The two modules are jointly optimized to detect both locally- and globally-anomalous graphs. Extensive empirical results on 16 real-world graph datasets from various domains show that i) HimNet significantly outperforms the state-of-art methods and ii) it is robust to anomaly contamination. Codes are available at: \renewcommand\UrlFont{\color{blue}\tt}\url{https://github.com/Niuchx/HimNet} 

\end{abstract}

\section{Introduction}

Graphs are widely used to model complex relationships between data instances in various fields, such as social networks, bioinformatics, chemistry, etc. Graph neural networks (GNNs) have become the predominant approach to learning effective node/graph representations and have achieved impressive performance in many graph-related tasks, such as node classification \cite{kipf2017semisupervised}, link prediction \cite{zhang2018link} and graph classification \cite{xu2018how}. Despite the remarkable success achieved by GNNs, it is still challenging for GNNs to tackle some notoriously difficult tasks. Graph-level anomaly detection (GLAD), which aims to identify abnormal graphs that exhibit deviant structures and node attributes in comparison to the majority in a set of graphs, is one of such tasks.

In recent years, a number of graph anomaly detection methods have been proposed. However, a majority of them focus on the detection of abnormal nodes or edges in a single graph \cite{ding2021inductive,kumagai2021semi,jin2021anemone,liu2021anomaly,wang2022cross,qiao2023truncated}. In contrast, graph-level anomaly detection is significantly less explored, despite its great importance and broad application \cite{ma2022deep,aggarwal2010graph,lee2021descriptive}. In general, anomalous graphs can be any graphs that are abnormal in part or in whole, which are referred to as locally-anomalous or globally-anomalous graphs \cite{ma2022deep,luo2022deep}. The local abnormality requires a fine-grained inspection of the graphs, as it is primarily due to the presence of unusual local graph structures, e.g., nodes and their associated local neighborhoods, compared to the corresponding structures in the other graphs. The global abnormality, on the other hand, requires a holistic treatment of the graphs, as it is manifested only at the graph-level representations. Thus, the main challenge in GLAD is to learn normal patterns from both fine-grained and holistic views of graphs for identifying both locally- and globally-anomalous graphs.

A few GLAD methods have been introduced, e.g., \cite{ma2022deep,luo2022deep}. They employ knowledge distillation \cite{ma2022deep} or contrastive learning \cite{luo2022deep} on the node and graph representations to capture the local/global normal patterns. The key intuition of these methods is that the model trained to fit exclusively normal training graphs learns normality representations, on which abnormal test graphs would be discriminative from the normal graphs. Despite their effectiveness, the learned normality representations may not preserve the primary semantics of graph structures and attributes, since their learning objectives ignore these semantics and focus on enlarging the relative difference between normal and abnormal graphs in the representation space. Consequently, they become ineffective in detecting abnormal graphs in which semantic-rich graph representations are required. 

This paper introduces a novel approach, namely \underline{hi}erarchical \underline{m}emory \underline{net}works (\textbf{HimNet}), via a graph autoencoder architecture to learn hierarchical node and graph memory modules for GLAD, which not only help effectively differentiate normal and abnormal graphs but also preserve rich primary semantics. Autoencoder (AE) \cite{kingma2013auto,bengio2006greedy}, which utilizes a decoder to reconstruct the original input based on the representations learned by an encoder, is a widely-used approach to preserve the rich semantics of the input data in the new representation space. AE is also commonly used for anomaly detection in various domains \cite{zhai2016deep,chen2017outlier,zhou2017anomaly,zong2018deep,gong2019memorizing,park2020learning} since anomalies are generally difficult to reconstruct, and thus, they have a higher reconstruction error than normal samples. However, reconstructing graphs is difficult since it involves the reconstruction of diverse graph structures and attributes. Our hierarchical memory learning is designed to address this issue. Specifically, the node-level memory module captures the local normal patterns that describe the fine-grained, internal graph interactions among nodes, and it is optimized by minimizing \textit{a graph reconstruction error} between original input graphs and the graphs reconstructed from the node memory module. On the other hand, the graph-level memory module is dedicated to the learning of holistic normal patterns of graph-level representations, and it is optimized by minimizing \textit{a graph approximation error} between graph-level representations and their approximated representations based on the graph memory. The two modules are jointly optimized to detect both locally and globally anomalous graphs. Memory-augmented AEs \cite{gong2019memorizing,park2020learning} have been introduced to add a memory module for anomaly detection in image and video data. The memory module has shown promise in enabling improved detection performance. However, their memory module is not applicable to graph data. HimNet addresses this problem by learning hierarchical node and graph memory modules to capture the local and global normal patterns of those non-Euclidean graph data.

In summary, our main contributions include: \textbf{i}) we introduce a hierarchical node-to-graph memory network HimNet for GLAD, which is the first work of memory-based GLAD; \textbf{ii}) we introduce a three-dimensional node memory module that consists of multiple two-dimensional memory blocks (with each block capturing one type of normal pattern on the representations of all nodes), as well as a graph memory module with each memory block capturing graph-level normal patterns; and \textbf{iii}) we further propose to learn these two memory modules by jointly minimizing a graph reconstruction error and a graph approximation error. We evaluate the effectiveness of HimNet via extensive experiments on 16 GLAD datasets from different domains, which show that HimNet significantly outperforms several state-of-art models and it also demonstrates remarkable robustness to anomaly-contaminated training data.

\section{Related Work}
\subsection{Graph-level Anomaly Detection}

Graph anomaly detection has attracted increasing research interest in recent years and various methods have been proposed \cite{akoglu2015graph,pang2021deep}. However, most of them focus on detecting anomalous nodes or edges in a single graph \cite{ding2021inductive,kumagai2021semi,jin2021anemone,liu2021anomaly,wang2022cross,qiao2023truncated}; significantly fewer studies are conducted on GLAD. Recently, a few GLAD methods have been proposed. These works can be divided into two categories: two-step methods and end-to-end methods. The first category typically obtains graph representations using graph kernels (e.g., Weisfeiler-Leman Kernel \cite{shervashidze2011weisfeiler} and propagation kernels \cite{neumann2016propagation}), or advanced GNNs (such as Graph2Vec \cite{narayanan2017graph2vec} and InfoGraph \cite{sun2020infograph}). An off-the-shelf anomaly detector is then applied to the learned graph representations to detect abnormal graphs, such as $k$-nearest-neighbor distance \cite{pang2015lesinn}, isolation forest \cite{liu2008isolation}, local outlier factor \cite{breunig2000lof}, and one-class support vector machine \cite{scholkopf1999support}. However, the two-step methods may achieve suboptimal performance since the anomaly detectors are independent of the graph representation learning. To address this issue, end-to-end methods unify graph representation learning and anomaly detection. Typically, they utilize powerful GNNs as the backbone and learn graph representations tailored for graph anomaly detection. For example, \cite{zhao2021using} applied the Deep SVDD objective \cite{ruff2018deep} on top of the GNN-based graph representations for anomalous graph detection. \cite{ma2022deep} utilized random knowledge distillation on both node and graph representations to capture graph regularity information. Some works also employed contrastive learning strategy for detecting anomalous graphs \cite{luo2022deep,liu2023good,qiu2022raising}. These methods show better performance than the two-step methods, but they focus on learning discriminative representations only, which may fail to preserve the primary graph semantics. Our method addresses this issue by learning hierarchical memory modules with the objective of preserving as much semantic as possible in the representation space.

\subsection{Memory Networks}

Due to the ability to store and retrieve important information, memory networks have been proposed and successfully applied to a wide range of domains
\cite{weston2015memory,li2016learning,kim2018memorization,wu2018unsupervised,gong2019memorizing,park2020learning}. For generative models, external memory is exploited to store local detail information \cite{li2016learning} and prevent the model collapsing problem \cite{kim2018memorization}. Considering that memory can be used to record prototypical patterns of normal data, a number of studies \cite{gong2019memorizing,park2020learning} proposed to augment AEs with a memory module for image or video anomaly detection. Despite the success of these methods, their memory networks are not applicable to GLAD as graph data is non-Euclidean and contains diverse graph structures and attributes where graph abnormality may exist. Our hierarchical node-to-graph memory modules are specifically designed to address this problem.

\section{Methodology}
\subsection{The GLAD Problem}

Let $\mathcal{G} = (\mathcal{V}, \mathcal{E})$ denote a graph, where $\mathcal{V}$ is the set of $N$ nodes and $\mathcal{E}$ is the set of edges. $\mathcal{E}$ is commonly represented by an adjacency matrix $\mathbf{A} \in [0, 1]^{N\times N}$ where $\mathbf{A}_{ij} = 1$ if node $i$ and $j$ are connected with an edge and $\mathbf{A}_{ij} = 0$ otherwise. If $\mathcal{G}$ is an attributed graph, the node features can be represented as $\mathbf{X} \in \mathbb{R}^{N\times d}$ where $d$ is the feature dimension. Therefore, a graph can also be denoted as $\mathcal{G} = (\mathbf{A},\mathbf{X})$. This work targets graph-level anomaly detection. Specifically, given a set of $K$ normal training graphs $\{\mathcal{G}_i=(\mathbf{A}_i, \mathbf{X}_i)\}_{i=1}^K$, we aim to learn an anomaly scoring function that assigns a high anomaly score to a test graph $\mathcal{G}$ if it significantly deviates from the majority in a set of graphs. 

\begin{figure}[!ht]
    \centering
    \includegraphics[width=0.96\textwidth]{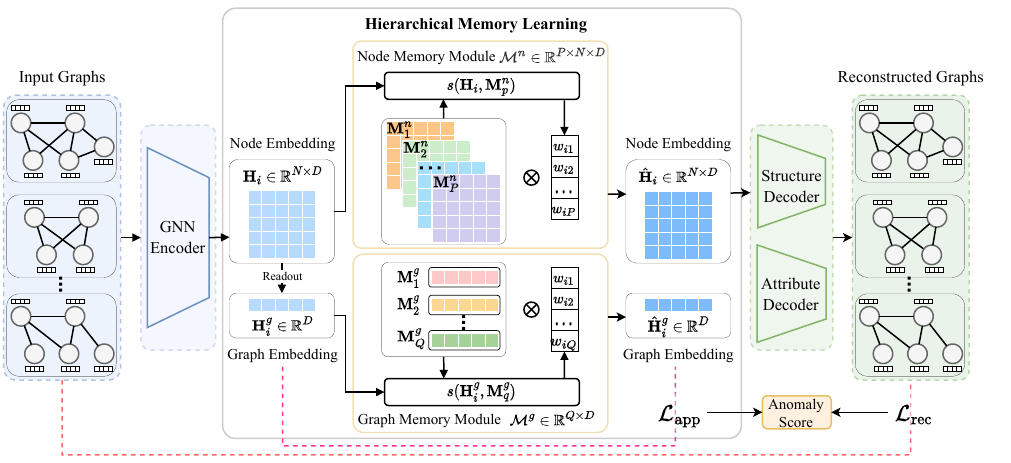}
    \caption{Overview of the proposed HimNet. It learns a three-dimensional node memory module $\mathcal{M}^{n} \in \mathbb{R}^{P\times N \times D}$ and a two-dimensional graph memory module $\mathcal{M}^{g}\in \mathbb{R}^{Q \times D}$, where $P$ and $Q$ denotes the number of node and graph memory blocks respectively, $D$ is the dimensionality of learned node representations, and $N$ is the number of nodes.} 
    \label{frame}
\end{figure}

\subsection{Overview of the Proposed Hierarchical Memory Networks}
We introduce HimNet to learn hierarchical node and graph memory blocks that respectively capture local and global normal patterns for GLAD. HimNet consists of four key components, namely graph encoder, graph decoder, node and graph memory modules, as shown in Fig. \ref{frame}. The node memory module is designed as a three-dimensional tensor that consists of multiple two-dimensional memory blocks, with each block capturing one type of normal pattern on all nodes. On the other hand, the graph memory is designed as a two-dimensional matrix, with each memory block capturing normal patterns on graph-level representations. These two memory modules are trained to capture hierarchical normal patterns of graph data, enabling the detection of locally- and globally-anomalous graphs.

Given an input graph, the graph encoder learns the node-level representation, and graph-level representation is obtained by applying a readout function on it. Traditionally, the graph decoder takes the node-level representation as input to reconstruct the input graph. However, this would increase the probability of the graph autoencoder reconstructing the abnormal graphs well. To tackle this issue, HimNet decouples the decoder from the encoder by replacing the encoded node-level representation with a combination of local patterns in the node memory module. Moreover, the graph-level representation is approximated by global patterns in the graph memory module. Then, the proposed model is optimized by minimizing graph reconstruction error and graph approximation error. This not only optimizes the parameters of the encoder and decoder but also forces the two memory modules to learn prime patterns of normal training graphs at both node and graph levels. After model optimization, given a test graph, the decoder takes the local normal patterns in the node memory module as input and the graph-level representation is approximated by global normal patterns in the graph memory module. In this way, the graph reconstruction error together with the graph approximation error can be used as an effective anomaly score.

\subsection{Graph Autoencoder}
In this paper, we build HimNet using a graph autoencoder (GAE) \cite{kipf2016variational} architecture to learn hierarchical memory modules. Before delving into the details of HimNet, we give an introduction to graph autoencoder which consists of a GNN-based encoder and decoder. 

\subsubsection{Encoder}
GNNs have recently emerged as a powerful class of deep-learning models for graph-structured data \cite{kipf2017semisupervised,veličković2018graph,xu2018how}. In this work, we employ GCN \cite{kipf2017semisupervised} as the graph encoder to generate the latent node-level and graph-level representations.

Let $\phi_e(\cdot:\Theta_e)$ be the encoder parameterized by $\Theta_e$. For every graph $\mathcal{G}_i = (\mathbf{A}_i, \mathbf{X}_i)$, the encoder takes the adjacency matrix $\mathbf{A}_i$ and node attributes $\mathbf{X}_i$ as input. The formulation of the encoder at $l$-th layer can be expressed as follows:
\begin{equation}
    \mathbf{H}^l_i = {\rm ReLU}(\hat{\mathbf{A}}_i\mathbf{H}^{(l-1)}_i\Theta_e^l)\,,
\end{equation}
where $\mathbf{H}^l_i$ and $\Theta_e^l$ represent the node representations and weight parameters of the GCN encoder at the $l$-th layer respectively, and ${\rm ReLU}(\cdot)$ is the non-linear activation function. $\hat{\mathbf{A}}_i = \tilde{\mathbf{D}}_i^{-\frac{1}{2}}\tilde{\mathbf{A}}_i\tilde{\mathbf{D}}_i^{-\frac{1}{2}}$, where $\tilde{\mathbf{A}}_i = \mathbf{A}_i + \mathbf{I}$ ($\mathbf{I}$ is an identity matrix) and $\tilde{\mathbf{D}}_i$ is the degree matrix of $\tilde{\mathbf{A}}_i$. $\mathbf{H}^{(l-1)}_i$ represents the node representation at the $(l-1)$-th layer and $\mathbf{H}^{(0)}_i = \mathbf{X}_i$. If the input graph $\mathcal{G}_i$ is a plain graph, the node degree is typically used as the attribute \cite{zhang2018end}. Assuming the output dimension of the encoder is $D$, the learned node representation can be formulated as $\mathbf{H}_i \in \mathbb{R}^{N \times D}$ where $N$ is the number of nodes in the graph.

To obtain the graph-level representation, a readout function is commonly applied to the learned node representation $\mathbf{H}_i$. There are many readout functions, such as maxing, averaging, summation, and concatenation \cite{wu2020comprehensive,zhang2020deep}. In this paper, we adopt the averaging function which calculates the mean of node representations along the node dimension to get the graph-level representation $\mathbf{H}^g_i \in \mathbb{R}^D$. The resulting representation $\mathbf{H}^g_i$ captures the overall structural and semantic information of the graph $\mathcal{G}_i$.

\subsubsection{Decoder}
To accurately reconstruct the original graph $\mathcal{G}_i$, two decoders $\phi_d^s(\mathbf{H}_i)$ and $\phi_d^a(\mathbf{H}_i)$, which take node representation $\mathbf{H}_i$ as input, are employed to reconstruct the graph structure and node attribute respectively.

For the graph structure decoder $\phi_d^s(\mathbf{H}_i)$, we implement it as the inner product of the latent node representation $\mathbf{H}_i$ as follows:
\begin{equation}
    \mathbf{A}_i^{'} = \sigma(\mathbf{H}_i\mathbf{H}_i^T)\,,
\end{equation}
where $\mathbf{A}_i^{'}$ denotes the reconstructed graph structure, $\mathbf{H}_i^T$ is the transpose of $\mathbf{H}_i$, and $\sigma(\cdot)$ represents the activation function.

To reconstruct the node attribute, we use the GCN \cite{kipf2017semisupervised} as the attribute decoder $\phi_d^a(\mathbf{H}_i)$ and the formulation at the $l$-th layer can be expressed as:
\begin{equation}
    \tilde{\mathbf{H}}_i^l = {\rm ReLU}(\hat{\mathbf{A}}_i\tilde{\mathbf{H}}_i^{(l-1)}\Theta_d^l)\,,
\end{equation}
where $\tilde{\mathbf{H}}_i^l$ and $\Theta_d^l$ represent the latent node representations and weight parameters at $l$-th layer of the decoder respectively, with $\tilde{\mathbf{H}}_i^{(0)} = \mathbf{H}_i$. We denote the reconstructed node attribute as $\mathbf{X}_i^{'}$, which is the output of the decoder $\phi_d^a(\mathbf{H}_i)$.

For each input graph $\mathcal{G}_i = (\mathbf{A}_i, \mathbf{X}_i)$, GAE is optimized to minimize the reconstruction errors on the graph structure and node attributes:
\begin{equation}\label{gae}
    \mathcal{L}_{\rm GAE} = \|\mathbf{A}_i - \mathbf{A}_i^{'}\|_F^2 + \|\mathbf{X}_i-\mathbf{X}_i^{'}\|_F^2\,,
\end{equation}
where $\|\cdot\|_F$ represents Frobenius norm. 

By minimizing Eq. (\ref{gae}), GAE is driven to fit the patterns of normal training graph data and preserve the semantics of them. During inference, GAE would produce higher reconstruction errors for anomalous graphs than normal graphs, as abnormal graphs are distinctive from normal graphs and are not accessible to GAE during the training process. Therefore, the reconstruction error $\mathcal{L}_{GAE}$ can be directly used as the criterion for anomaly detection. However, solely relying on $\mathcal{L}_{GAE}$ often cannot yield satisfactory anomaly detection performance, as demonstrated in the experiments section. This is primarily because graph is difficult to reconstruct, leading to less discriminative power in differentiating normal and abnormal graphs. Moreover, such a GAE cannot model graph-level patterns well, as graph representations are not explored in GAE. In this work, we propose to learn hierarchical memory modules to address this problem.

\subsection{Hierarchical Memory Learning}
Hierarchical memory learning consists of two memory modules: node and graph memory modules, which are designed to capture hierarchical node-to-graph patterns of the normal training graphs and facilitate the detection of graphs that are abnormal in part or in whole.

\subsubsection{Graph Memory Module}
The graph memory module aims to capture the prototypical patterns inherent in the graph representations $\{\mathbf{H}^g_i\}_{i=1}^K$ ($K$ is the number of training graphs) through a set of graph memory blocks, denoted as $\mathcal{M}^{g}=\{\mathbf{M}_q^g \in \mathbb{R}^{D}\}_{q=1}^Q$, where $Q$ is the total number of memory blocks and each  block $\mathbf{M}_q^g$ is of the same dimensionality size as the graph representation $\mathbf{H}_i^g$. 

Since the graph memory blocks capture prototypical patterns of graph representations, a graph representation $\mathbf{H}^g_i$ can be approximated using the following equation:
\begin{equation}\label{graph_app}
    \hat{\mathbf{H}}^g_i = \sum_{q=1}^Q w_{iq} \mathbf{M}_q^g, \ \ \ \ {\rm s.t.} \ \ \sum_{q=1}^Q w_{iq} = 1\,, 
\end{equation}
where $\hat{\mathbf{H}}^g_i$ is the approximated representation of $\mathbf{H}^g_i$ from the memory blocks, and $w_{iq}$ is the weight of the memory block $\mathbf{M}_q^g$ for $\mathbf{H}_i^g$, with the summation of the weights constrained to be one. The weight $w_{iq}$ reflects the correlation between each graph memory block and the graph representation, i.e., a higher correlation induces a larger weight. Therefore, to calculate $w_{iq}$, we first employ a cosine similarity function $s(\cdot)$ to measure the similarity between $\mathbf{M}_q^g$ and $\mathbf{H}_i^g$:
\begin{equation}
    s(\mathbf{H}_i^g, \mathbf{M}_q^g) = \frac{\mathbf{H}_i^g (\mathbf{M}_q^g)^T}{\|\mathbf{H}_i^g\|\|\mathbf{M}_q^g\|}\,.
\end{equation}
To impose the summation constraint, we further normalize the similarities via the following softmax operation to obtain the final weight: 
\begin{equation}
    w_{iq} = \frac{\exp(s(\mathbf{H}_i^g, \mathbf{M}_q^g))}{\sum_{q=1}^Q \exp(s(\mathbf{H}_i^g, \mathbf{M}_q^g))}\,.
\end{equation}

After obtaining the approximated graph representation $\hat{\mathbf{H}}_i^g$, we calculate the approximation error via the following $\mathcal{L}_{\rm app}$ loss:
\begin{equation}\label{graphmem}
    \mathcal{L}_{\rm app} = \|\mathbf{H}_i^g - \hat{\mathbf{H}}_i^g\|_F^2\,.
\end{equation}

In the training phase, the optimization of Eq. (\ref{graphmem}) not only minimizes the approximation error through an efficient combination of the graph memory blocks but also forces the graph memory blocks to learn the most crucial patterns of the graph representations. In this way, during the test phase, the approximation errors for normal and abnormal graphs would become distinct. This occurs because the approximated graph representation is constructed solely through the weighted combination of the learned normal patterns of graph representations. 

\subsubsection{Node Memory Module}

Different from the graph memory module that captures the normal patterns at the graph-level representations, the node memory module is designed to capture the fine-grained, normal patterns on the node representations $\{\mathbf{H}_i \in \mathbb{R}^{N\times D}\}_{i=1}^K$. Specifically, the node memory module is designed as a three-dimensional tensor, consisting of $P$ two-dimensional memory matrices, $\mathcal{M}^{n}=\{\mathbf{M}^n_p \in \mathbb{R}^{N \times D}\}_{p=1}^{P}$, with each memory block $\mathbf{M}^n_p$ having the same dimensionality size as the representations of all nodes. This way helps effectively capture interactions across all nodes and their local neighborhood. 

To reduce the probability of the decoder reconstructing the abnormal graph unexpectedly, for a node representation $\mathbf{H}_i$, the node memory module approximates it with $\hat{\mathbf{H}}_i$ and feeds $\hat{\mathbf{H}}_i$ to the decoder. Formally, $\hat{\mathbf{H}}_i$ is obtained by:
\begin{equation}
    \hat{\mathbf{H}}_i = \sum_{p=1}^P w_{ip} \mathbf{M}_p^n, \ \ \ \ {\rm s.t.} \ \ \sum_{p=1}^P w_{ip} = 1\,,
\end{equation}
where $w_{ip}$ is the weight of the memory block $\mathbf{M}_p^n$ for $\mathbf{H}_i$ and the summation of $w_{ip}$ is constrained to 1. To compute the value of $w_{ip}$, we adopt the same approach used in the graph memory block. Specifically, we first calculate the similarity between the node representation $\mathbf{H}_i$ and each node memory block $\mathbf{M}_p^n$. Then, we normalize the similarities through the softmax function to obtain the final weight value $w_{ip}$.

The approximated node representation $\hat{\mathbf{H}}_i$ is fed as the input to the graph structure decoder $\phi_d^s(\cdot)$ and node attribute decoder $\phi_d^a(\cdot)$. In this way, the reconstruction error based on $\hat{\mathbf{H}}_i$ can be reformulated as:
\begin{equation}\label{nodemem}
    \mathcal{L}_{\rm rec} = \|\mathbf{A}_i - \phi_d^s(\hat{\mathbf{H}}_i)\|_F^2 + \|\mathbf{X}_i-\phi_d^a(\hat{\mathbf{H}}_i)\|_F^2\,.
\end{equation}

Compared to GAE which reconstructs the original graph depending on the encoded node representation, the node memory module performs graph construction solely based on the weighted combination of the node memory blocks. During training, the graph memory blocks are driven to learn the most representative patterns in the encoded node representations by minimizing the reconstruction error in Eq. (\ref{nodemem}). While in the testing phase, regardless of whether the input graph is normal or not, the decoder only takes different combinations of the learned normal patterns as input and outputs the normal-like graphs. Consequently, the reconstruction errors between normal and abnormal graphs would become significantly different. Overall, the node memory module decouples the decoder from the encoder, resulting in the graph reconstruction being more sensitive to the anomaly. 

\subsection{Training and Inference}
\subsubsection{Training Objective}
By jointly employing the graph and node memory modules, HimNet aims to capture the hierarchical normal patterns of graphs. To achieve this goal, for each graph, our model is optimized by minimizing the combined objective of Eq. (\ref{graphmem}) and Eq. (\ref{nodemem}):
\begin{equation}\label{recall}
    \mathcal{L}_{\rm rec}^{'} = \|\mathbf{A}_i - \phi_d^s(\hat{\mathbf{H}}_i)\|_F^2 + \|\mathbf{X}_i-\phi_d^a(\hat{\mathbf{H}}_i)\|_F^2 + \|\mathbf{H}_i^g - \hat{\mathbf{H}}_i^g\|_F^2\,.
\end{equation}

To further enhance the discrimination of HimNet for normal and abnormal graphs, we adopt the hard shrinkage strategy \cite{gong2019memorizing} to promote the sparsity of weight parameters $w_{ip}$ and $w_{iq}$. Besides, the entropy of $w_{ip}$ and $w_{iq}$ are calculated and minimized during the training, which can be formulated as follows:
\begin{equation}\label{entropy}
    \mathcal{L}_{\rm entropy} = \sum_{i=1}^{P} -w_{ip}\log w_{ip} + \sum_{i=1}^{Q} -w_{iq}\log w_{iq}\,.
\end{equation}
By employing the hard shrinkage and the entropy term, the weight parameters would become more sparse, i.e., the encoded node and graph representations are approximated with fewer memory blocks. This requires the chosen memory blocks to be more relevant to the encoded representations and also forces the memory blocks to learn more informative patterns.

The final training objective is obtained by combining Eq. (\ref{recall}) and Eq. (\ref{entropy}):
\begin{equation}\label{obj}
    \mathcal{L}_{\rm train} = \mathcal{L}_{\rm rec}^{'} + \alpha\mathcal{L}_{\rm entropy}\,,
\end{equation}
where $\alpha$ is a hyperparameter controlling the importance of the entropy term.

\subsubsection{Inference}
By optimizing Eq. (\ref{obj}), HimNet can capture the hierarchical normal patterns of graphs. As a result, for a normal graph, HimNet is capable of reconstructing it effectively with the memory blocks learned in both node and graph memory modules. However, for an abnormal graph, the value in Eq. (\ref{recall}) tends to be high. Therefore, we adopt the loss term Eq. (\ref{recall}) as the anomaly score, where a higher value indicates a larger probability of being an abnormal graph.

\section{Experiments}

\subsection{Experimental Setups}
\subsubsection{Datasets}
To verify the effectiveness of HimNet, we conduct experiments on 16 publicly available graph datasets from two popular application domains: i) biochemical molecules (PROTEINS\_full, ENZYMES, AIDS, DHFR, BZR, COX2, DD, NCI1, HSE, MMP, p53, PPAR-gamma, and hERG) and ii) social networks (IMDB, REDDIT, and COLLAB). The statistics of these graph datasets\footnote{All the graph datasets are available on \url{https://chrsmrrs.github.io/datasets/docs/datasets/} except hERG which is obtained from \url{https://tdcommons.ai/single_pred_tasks/tox/}} are summarized in Table \ref{statis}. Specifically, the first six datasets in Table \ref{statis} are attributed graphs and the other datasets consist of plain graphs. Moreover, HSE, MMP, p53, and PPAR-gamma contain real anomalies while the other graph datasets are originally constructed for graph classification. Following \cite{ma2022deep,pang2019deep,campos2016evaluation,liu2008isolation}, these datasets are converted for GLAD by treating the minority class in these datasets as anomalies. 

\begin{table*}[ht]
\centering
\caption{Key Statistics of Graph Datasets.}
\label{statis}
\setlength{\tabcolsep}{1.6mm}
\resizebox{0.95\textwidth}{!}{
\begin{tabular}{lccccc}
\hline
\textbf{Dataset} & \textbf{Category} &\textbf{\# Graphs} & \textbf{\# Avg.Nodes} & \textbf{\# Avg.Edges} & \textbf{\# Anomaly Rate} \\
\hline
PROTEINS$\_$full & Biochemical Molecules & 1,113 & 39.06 & 72.82  &0.60\\
ENZYMES & Biochemical Molecules & 600 & 32.63 & 62.14 &0.17\\
AIDS & Biochemical Molecules & 2,000 & 15.69 & 16.2 &0.20\\
DHFR & Biochemical Molecules & 467 & 42.43 & 44.54  &0.61\\
BZR & Biochemical Molecules & 405 & 35.75 & 38.36 &0.79\\
COX2 & Biochemical Molecules & 467 & 41.22 & 43.45 &0.78\\
DD & Biochemical Molecules & 1,178 & 284.32 & 715.66 &0.58\\
NCI1 & Biochemical Molecules & 4,110 & 29.87 & 32.3 &0.50\\
HSE & Biochemical Molecules & 8,417 & 16.89 & 17.23 &0.04\\
MMP & Biochemical Molecules & 7,558  & 17.62 & 17.98 &0.16\\
p53 & Biochemical Molecules & 8,903 & 17.92 & 18.34  &0.10\\
PPAR-gamma & Biochemical Molecules & 8,451 & 17.38 & 17.72 &0.06\\
hERG & Biochemical Molecules & 655 & 26.48 & 28.79 &0.31\\
IMDB & Social Networks & 1,000 & 19.77 & 96.53  &0.50\\
REDDIT & Social Networks & 2,000 & 429.63 & 497.75 &0.50\\
COLLAB & Social Networks & 5,000 & 74.49 & 2,457.78 &0.52\\
\hline
\end{tabular}
}
\end{table*}

\subsubsection{Competing Methods}
Several competing methods from two categories are used for comparison to the proposed method. The first category consists of two-step methods that use state-of-art graph representation learning methods to extract graph representations and then apply an advanced anomaly detector on the learned representations to identify anomalous graphs. Specifically, we employ InfoGraph \cite{sun2020infograph}, Weisfeiler-Lehamn (WL) \cite{shervashidze2011weisfeiler}, and propagation kernel (PK) \cite{neumann2016propagation} as the graph encoder respectively and utilize isolation forest as the anomaly detector following \cite{ma2022deep}. The second category of baselines includes OCGCN \cite{zhao2021using}, GLocalKD \cite{ma2022deep}, and GAE \cite{kipf2016variational} that are trained in an end-to-end manner. OCGCN \cite{zhao2021using} applied an SVDD objective on top of GCN-based representation for graph anomaly detection. GLocalKD \cite{ma2022deep} utilized random knowledge distillation to identify anomalies. GAE \cite{kipf2016variational} used the graph reconstruction error to detect anomalous graphs.

\subsubsection{Implementation Details}
To ensure fair comparisons, we utilize a three-layer GCN \cite{kipf2017semisupervised} as the graph encoder following \cite{ma2022deep}. The dimensions of the latent layer and output feature are set to 512 and 256 respectively. The node attribute decoder is a two-layer GCN with the dimension of the latent layer set as 256. The batch size is 300 for all datasets except for HSE, MMP, p53, and PPAR-gamma whose bath size is 2000 since these datasets contain more graphs. The hyperparameter $\alpha$ is set to 0.01 for all datasets. This work targets detecting anomalous graphs within multiple graphs. However, the number of nodes varies across graphs which hinders the parallel processing of graph data. To address this issue, we augment the adjacency and the attribute matrices with zero padding to match the same size of the largest graph.

\subsubsection{Evaluation}
We employ the commonly used metric, area under receiver operating characteristic curve (AUC), to evaluate the anomaly detection performance. A higher AUC value indicates better performance. The mean and standard deviation of AUC results are reported by performing 5-fold cross-validation for all datasets except for HSE, MMP, p53 and PPAR-gamma which have widely used predefined train and test splits. For these datasets, we report the results by running the experiments five times with different random seeds.

\subsection{Comparison to State-of-the-art Methods}
The AUC results of the proposed method and the competing methods are reported in Table \ref{auroc}. From the average rank results in the table, we can see end-to-end methods generally perform better than two-step methods, which highlights the significance of learning tailored representations for graph-level anomaly detection. Further, our method outperforms all the methods on 13 out of 16 datasets and achieves highly competitive performance in the remaining three datasets. In comparison to GAE \cite{kipf2016variational}, our method incorporates two memory modules to learn hierarchical normal patterns. The performance improvements over GAE \cite{kipf2016variational} and other counterparts demonstrate the effectiveness of exploiting memory modules to capture hierarchical normal patterns for anomalous graph detection. Note that GAE performs poorly on NCI1 and REDDIT while our method achieves very promising results. This demonstrates that the rich semantics learned in HimNet allow significantly better performance than GAE in distinguishing abnormal and normal graphs, especially when the graphs are large and difficult to reconstruct, e.g., those in REDDIT.

We also perform a paired Wilcoxon signed rank test \cite{woolson2007wilcoxon} to verify the statistical significance of HimNet against the baselines across all 16 datasets and the results are shown in the bottom line of Table \ref{auroc}. We can see that our method surpasses all baseline approaches with a confidence level greater than 98\%. 

\begin{table*}[ht]
\centering
\caption{AUC results (in percent, mean$\pm$std) on 16 real-world graph datasets. The best and second performances in each row are boldfaced and underlined respectively.}
\label{auroc}
\setlength{\tabcolsep}{1.6mm}
\resizebox{0.95\textwidth}{!}{
\begin{tabular}{l|cccccc|c}
\hline
\textbf{Dataset} & \textbf{InfoGraph-iF} &\textbf{WL-iF} & \textbf{PK-iF} & \textbf{OCGCN} & \textbf{GLocalKD} & \textbf{GAE} & \textbf{Our}\\
\hline
PROTEINS$\_$full  & 46.4$\pm$1.9 & 63.9$\pm$1.8 & 62.7$\pm$0.9 & 71.8$\pm$3.6 & \textbf{78.5}$\pm$3.4 & 76.6$\pm$2.2& \underline{77.2$\pm$1.5}\\
ENZYMES  &  48.3$\pm$2.7 & 49.8$\pm$2.9 & 49.3$\pm$1.3 & \underline{61.3$\pm$8.7} & \textbf{63.6}$\pm$6.1 & 51.6$\pm$2.7 & 58.9$\pm$7.6\\
AIDS  & 70.3$\pm$3.6 & 63.2$\pm$5.0 & 47.6$\pm$1.4 & 66.4$\pm$8.0 & \underline{99.2$\pm$0.4} & 99.0$\pm$0.5& \textbf{99.7$\pm$0.3}\\
DHFR  & 48.9$\pm$1.5 & 46.6$\pm$1.3 & 46.7$\pm$1.3 & 49.5$\pm$8.0 & \underline{55.8$\pm$3.0} & 51.4$\pm$3.6& \textbf{70.1$\pm$1.7}\\
BZR & 52.8$\pm$6.0 & 53.3$\pm$3.2 & 52.5$\pm$5.2 & 65.8$\pm$7.1 & \underline{67.9$\pm$6.5} &65.5$\pm$8.3 & \textbf{70.3$\pm$5.4}\\
COX2 & 58.0$\pm$5.2 & 53.2$\pm$2.7 & 51.5$\pm$3.6 & \underline{62.8$\pm$7.2} & 58.9$\pm$4.5 & 58.7$\pm$4.9& \textbf{63.7$\pm$7.6}\\
DD & 47.5$\pm$1.2 & 69.9$\pm$0.6 & 70.6$\pm$1.0 & 60.5$\pm$8.6 & \underline{80.5$\pm$1.7} & 80.4$\pm$1.7 & \textbf{80.6$\pm$2.1}\\
NCI1  & 49.4$\pm$0.9 & 54.5$\pm$0.8 & 53.2$\pm$0.6 & 62.7$\pm$1.5 & \underline{68.3$\pm$1.5} &35.0$\pm$1.9 & \textbf{68.6$\pm$1.9}\\
HSE  & 48.4$\pm$2.6 & 47.7$\pm$0.0 & 48.9$\pm$0.3 & 38.8$\pm$4.1 & \underline{59.1$\pm$0.1} & 59.0$\pm$0.5& \textbf{61.3$\pm$3.9}\\
MMP  & 53.9$\pm$2.2 & 47.5$\pm$0.0 & 48.8$\pm$0.2 & 45.7$\pm$3.8 &  \underline{67.6$\pm$0.1} & 67.3$\pm$0.4& \textbf{70.3$\pm$2.9}\\
p53  & 51.1$\pm$1.4 & 47.3$\pm$0.0 & 48.6$\pm$0.4 & 48.3$\pm$1.7 & 63.9$\pm$0.2 & \underline{64.0$\pm$0.1}& \textbf{64.6$\pm$0.2}\\
PPAR-gamma & 52.1$\pm$2.3 & 51.0$\pm$0.0 & 49.9$\pm$1.7 & 43.1$\pm$4.3 & 64.4$\pm$0.1 & \underline{65.8$\pm$1.6}& \textbf{71.1$\pm$3.4}\\
hERG & 60.7$\pm$3.3 & 66.5$\pm$4.2 & 67.9$\pm$3.4 & 56.9$\pm$4.9 & \underline{70.4$\pm$4.9} & 68.1$\pm$8.7& \textbf{75.4$\pm$3.2}\\
IMDB  & 52.0$\pm$2.8 & 44.2$\pm$3.2 & 44.2$\pm$3.5 & 53.6$\pm$14.8 & 51.4$\pm$3.9 & \underline{65.2$\pm$4.4}& \textbf{68.3$\pm$3.2}\\
REDDIT & 45.7$\pm$0.3 & 45.0$\pm$1.3 & 45.0$\pm$1.2 & 75.9$\pm$5.6 & \textbf{78.2}$\pm$1.6 &21.8$\pm$1.9 & \underline{78.0$\pm$2.5}\\
COLLAB  & 45.3$\pm$0.3 & 50.6$\pm$2.0 & 52.9$\pm$2.3 & 40.1$\pm$18.3 & 52.5$\pm$1.4 &\underline{52.8$\pm$1.4} & \textbf{55.3$\pm$3.2}\\

\hline
Avg.Rank &5.25 &5.63 &5.31 &4.75 &2.31 &3.50 &1.25 \\
p-value &0.0004 &0.0004 &0.0004 &0.0008 &0.0113 &0.0004 & - \\
\hline
\end{tabular}
}
\end{table*}

\subsection{Robustness w.r.t Anomaly Contamination}
\label{anomalycon}
This subsection evaluates the robustness of HimNet under different levels of anomaly contamination in training data. This scenario is generally very realistic since the graph data collected in the world may be contaminated by anomalies and noises. To simulate this setting, given the original training data that contain normal and abnormal data, instead of discarding abnormal data, we combine $\tau\%$ of the abnormal data with the normal training data to form the contaminated training data. Specifically, we vary the anomaly contamination rate $\tau$ from $0\%$ to $16\%$ and compare the performance of HimNet, with the two best competing methods -- GLocalKD \cite{ma2022deep} and GAE \cite{kipf2016variational} -- as the baselines. Without loss of generality and due to the page limits, we perform experiments on four datasets, including three from biochemical molecules (AIDS, BZR, and DHFR) and one from social networks (IMDB).

\begin{figure}[htbp]
    \centering
    \subfigure[AIDS]{\includegraphics[width=0.23\textwidth]{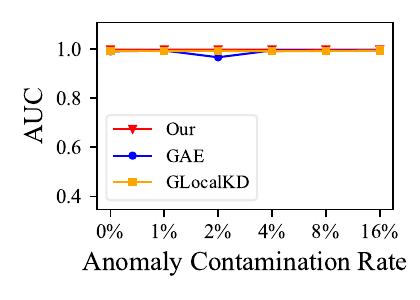}}
    \subfigure[BZR]{\includegraphics[width=0.23\textwidth]{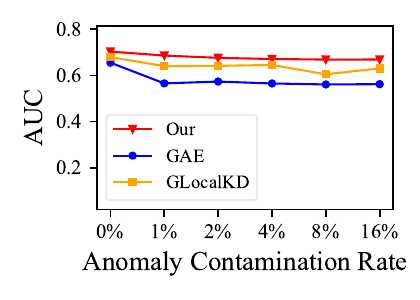}}
    \subfigure[DHFR]{\includegraphics[width=0.23\textwidth]{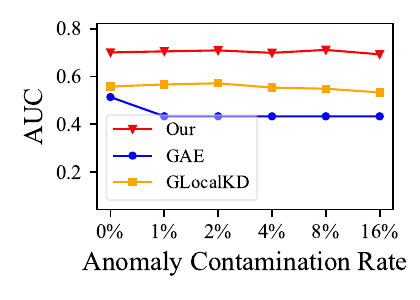}}
    \subfigure[IMDB]{\includegraphics[width=0.23\textwidth]{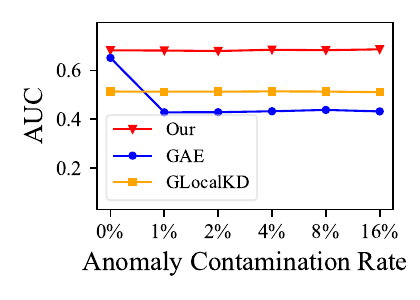}}
    \caption{Results of GAE, GLocalKD, and HimNet under various contamination rates.}
    \label{performancewithanomaly}
\end{figure}

Fig. \ref{performancewithanomaly} shows the AUC results of GAE, GLocalKD, and HimNet w.r.t. different anomaly contamination rates. Compared to the two baselines, our method achieves the best anomaly detection performance in all settings, particularly on DHFR and IMDB. We can also see that both GLocalKD and our method demonstrate consistent performance on all datasets for different anomaly contamination rates, while GAE experiences a significant performance drop with a slight increase in the number of anomalous training instances, except for the AIDS dataset. The reason for the superior and stable performance of our method is its ability to learn and store hierarchical patterns of the majority of training data. As a result, anomalous graphs can be readily detected since they cannot be reconstructed effectively using the learned hierarchical memory blocks. Note that all three methods perform similarly on the AIDS dataset, which could be due to the distinguishability between normality and abnormality being more apparent compared to the other datasets.

\subsection{Ablation Study}

We use the GAE as our base model to evaluate the importance of our proposed node and graph memory modules, which are the key driving components in HimNet. To verify the importance of each component, we conduct experiments on two variants of the proposed method, i.e., HimNet$_{\rm w/o\ node}$ and HimNet$_{\rm w/o\ graph}$ that respectively discard the node and graph memory module.

The results of HimNet, its two variants, and GAE are reported in Table \ref{ablation}. From the table, we can derive the following observations. First, by incorporating node or graph memory module into GAE, the anomaly detection performance is significantly enhanced on nearly all datasets, which verifies the effectiveness of each of our proposed memory modules. Using GAE without memory modules are ineffective on some challenging datasets with large graphs and/or complex node attributes, such as DHFR and REDDIT. Second, HimNet$_{\rm w/o\ node}$ and HimNet$_{\rm w/o\ graph}$ perform differently across graph datasets, which indicates that the dominance of locally or globally anomalous graphs varies across the graph datasets. For example, HimNet$_{\rm w/o\ graph}$ outperforms HimNet$_{\rm w/o\ node}$ on NCI1, indicating the anomalous graphs are more dominated by locally anomalous graphs in NCI1. Third, the performance improvement over GAE is further boosted by the utilization of both node and graph memory modules. This demonstrates the importance of capturing the hierarchical normal patterns that enable the simultaneous detection of locally and globally anomalous graphs.

\begin{table*}[ht]
\centering
\caption{AUC performance of the proposed method and its variants.}
\label{ablation}
\setlength{\tabcolsep}{0.6mm}
\resizebox{0.53\textwidth}{!}{
\begin{tabular}{l|cccc}
\hline
\multirow{2}*{\textbf{Dataset}} & \multirow{2}*{\textbf{GAE} }&\multirow{2}*{\begin{tabular}[c]{@{}c@{}}\textbf{HimNet}\\ \textbf{w/o node}\end{tabular}} & \multirow{2}*{\begin{tabular}[c]{@{}c@{}}\textbf{HimNet}\\ \textbf{w/o graph}\end{tabular}} & \multirow{2}*{\textbf{HimNet}} \\
\\
\hline
PROTEINS$\_$full&76.6$\pm$2.2 &76.4$\pm$2.5 &76.3$\pm$3.1 & \textbf{77.2$\pm$1.5}  \\
ENZYMES         &51.6$\pm$2.7 &52.0$\pm$4.3 &55.7$\pm$5.4 & \textbf{58.9$\pm$7.6} \\
AIDS            &99.0$\pm$0.5 &99.3$\pm$0.4 &99.6$\pm$0.3 & \textbf{99.7$\pm$0.3}  \\
DHFR            &51.4$\pm$3.6 &51.9$\pm$3.5 &68.3$\pm$6.7 & \textbf{70.1$\pm$1.7}  \\
BZR             &65.5$\pm$8.3 &69.9$\pm$4.9 &67.4$\pm$4.8 & \textbf{70.3$\pm$5.4}  \\
COX2            &58.7$\pm$4.9 &59.7$\pm$7.0 &\textbf{64.7$\pm$5.9} & 63.7$\pm$7.6 \\
DD              &80.4$\pm$1.7 &80.2$\pm$1.8 &80.4$\pm$2.0 & \textbf{80.6$\pm$2.1}  \\
NCI1            &35.0$\pm$1.9 &33.9$\pm$5.7 &62.2$\pm$2.4 & \textbf{68.6$\pm$1.9}  \\
HSE             &59.0$\pm$0.5 &59.4$\pm$0.4 &61.6$\pm$4.3 & \textbf{61.3$\pm$3.9} \\
MMP             &67.3$\pm$0.4 &68.7$\pm$0.8 &69.1$\pm$0.2 & \textbf{70.3$\pm$2.9} \\
p53             &64.0$\pm$0.1 &64.4$\pm$0.1 &\textbf{64.9$\pm$0.8} & 64.6$\pm$0.1 \\
PPAR-gamma      &65.8$\pm$1.6 &67.6$\pm$0.1 &67.5$\pm$3.9 & \textbf{71.1$\pm$3.4}  \\
hERG            &68.1$\pm$8.7 &68.7$\pm$0.8 &73.2$\pm$3.2 & \textbf{75.4$\pm$3.2}   \\
IMDB            &65.2$\pm$4.4 &65.2$\pm$4.4 &66.0$\pm$3.0 & \textbf{68.3$\pm$3.2}  \\
REDDIT          &21.8$\pm$1.9 &21.9$\pm$2.4 &31.1$\pm$8.9 & \textbf{78.0$\pm$2.5}  \\
COLLAB          &52.8$\pm$1.4 & 52.2$\pm$1.6 & 51.7$\pm$1.6 &\textbf{55.3$\pm$3.2}  \\
\hline
\end{tabular}
}
\end{table*}

\subsection{Analysis of Hyperparameters}
\begin{figure}[htbp]
    \centering
    \subfigure
    {\includegraphics[width=0.44\textwidth]{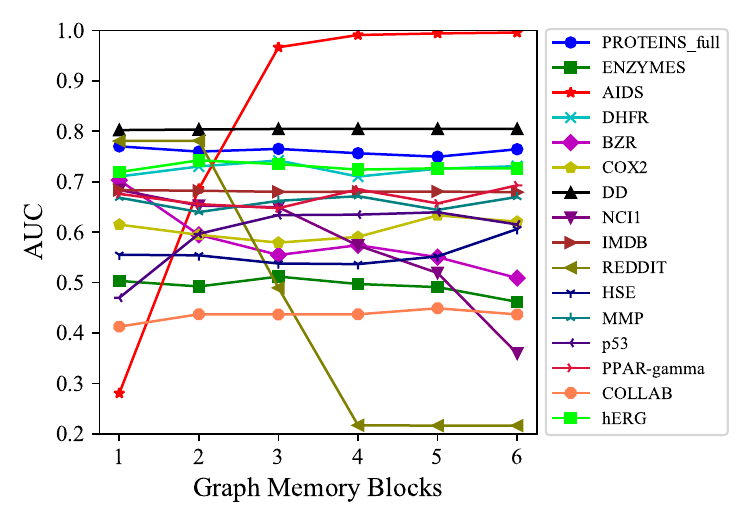}}
    \subfigure
    {\includegraphics[width=0.44\textwidth]{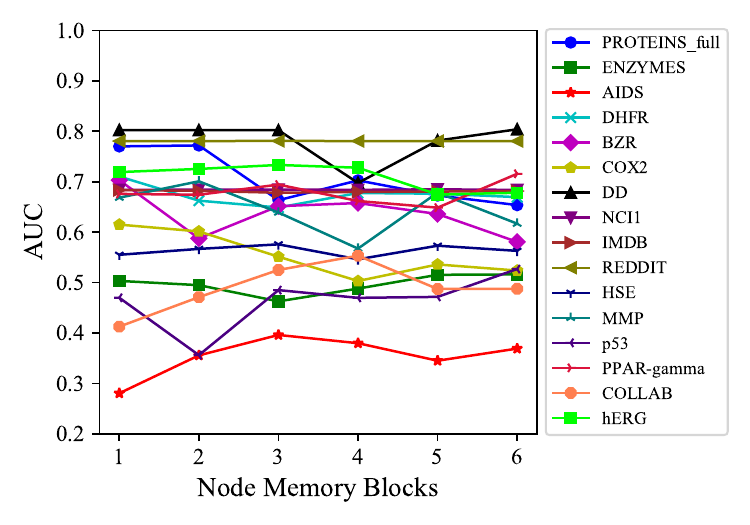}}
    \caption{Results of HimNet w.r.t different numbers of graph and node memory blocks.}
    \label{memnum}
\end{figure}

We examine the sensitivity of our method HimNet w.r.t the number of memory blocks in node and graph memory modules. Specifically, for one memory module, we fix the number of memory blocks to one and vary the number of memory blocks in the other memory module across $\{1,2,3,4,5,6\}$. The results of all graph datasets are reported in Fig. \ref{memnum}. The results show that even using one memory block in each memory module, the proposed method can still achieve promising performance on some datasets, such as DD, DHFR, REDDIT, and hERG. This may be because the normal graphs in these datasets are more homogeneous and deviate from the abnormal graphs distinctly. HimNet is generally more robust to the numbers of graph memory blocks than the number of node memory blocks except on the AIDS, BZR, NCI1, and REDDIT. Also, increasing the number of memory blocks does not always bring better results. In some cases, it can even degrade the detection performance. This is mainly because the larger memory modules may boost the expressiveness of memory modules, leading to the failure cases that the abnormal graphs can also be well reconstructed. 

\section{Conclusion}
This paper proposes hierarchical memory networks (HimNet) to learn hierarchical node and graph memory modules for graph-level anomaly detection. The node and graph memory module explicitly capture the hierarchical normal patterns of graphs by jointly minimizing graph reconstruction and graph approximation errors. The learned hierarchical memory blocks enable effective detection of both locally- and globally-anomalous graphs. Extensive experiments demonstrate the superiority of HimNet in detecting anomalous graphs compared to state-of-the-art methods. Furthermore, HimNet achieves promising performance even when the training data is largely contaminated by anomaly graphs, which shows its applicability in real-world applications with unclean training data. 

\noindent \textbf{Acknowledgement.} This work is partially supported by Australian Research Council under Grant DP210101347.

\newpage
\noindent\textbf{Ethical Statement:} In this work, we study the problem of graph-level anomaly detection which aims to identify abnormal graphs that exhibit unusual patterns in comparison to the majority in a graph set. Since graphs are widely used in various domains, anomaly detection on graphs has broad applications, such as identifying toxic molecules from chemical compound graphs and recognizing abnormal internet activity graphs. To capture the hierarchical normal patterns of graph data, we propose hierarchical memory networks to learn node and graph memory modules. The proposed method enables the detection of both locally and globally anomalous graphs. For all the used data sets in this paper, there is no private personally identifiable information or offensive content. However, when using the proposed method for solving realistic problems, it is essential to ensure that appropriate measures are taken to protect the privacy of individuals. This may include anonymizing data, limiting access to sensitive information, or obtaining informed consent from individuals before collecting their data. 

\bibliographystyle{splncs04}
\bibliography{reference}

\end{document}